\documentclass[conference]{IEEEtran}

\usepackage{cite}
\usepackage[T1]{fontenc}
\usepackage{amsmath,amsfonts}
\usepackage{graphicx}
\usepackage{booktabs}
\usepackage{xcolor}
\usepackage{url}
\usepackage{array}
\usepackage{hyperref}
\usepackage{eso-pic}

\hypersetup{hidelinks}
\title{A Machine Learning Framework for Predicting Restaurant
Food Waste to Support Sustainable Food Management}

\author{
\IEEEauthorblockN{
Md Mehedi Hasan Naeem, Md Ashraful Islam, Moumita Barua,\\[-0.2ex]
Ishtiyak Ahmmad Araf and Md. Arefin Haque Mahir
}
\IEEEauthorblockA{
{\footnotesize Department of Computer Science and Engineering,
Jatiya Kabi Kazi Nazrul Islam University,
Trishal, Mymensingh-2224, Bangladesh}\\[-0.2ex]
{\scriptsize
\{mehedinaeem00, ashrafulofficialc, moumitabarua.jkkniucse,
ishtiyakahmmad1149, arefinmahir2026\}@gmail.com}
}
}

\newcommand{\SPICSCONFirstPageNotices}{
\AddToShipoutPictureFG*{
  \AtPageUpperLeft{
    \raisebox{-0.26in}[0pt][0pt]{
      \hspace{0.55in}
      \parbox[t]{7.40in}{
      \raggedright
      \fontsize{8}{9}\selectfont
      2026 IEEE International Conference on Signal Processing,
      Information, Communication and Systems (SPICSCON)\\
      13-14 August 2026, Faculty of ECE, Bangladesh Army University
      of Engineering \& Technology (BAUET) , Qadirabad,
      Natore-6431, Bangladesh
}
    }
  }
  \AtPageLowerLeft{
    \raisebox{0.34in}[0pt][0pt]{
      \hspace{0.55in}
      \fontsize{7}{8}\selectfont
      979-8-3195-1270-3/26/\$31.00 \textcopyright 2026 IEEE
    }
  }
}
}

\begin{document}

\SPICSCONFirstPageNotices
\maketitle


\begin{abstract}
Food waste in the restaurant sector poses a substantial challenge
to environmental sustainability and economic efficiency. This paper
presents an exploratory machine learning framework for estimating
daily restaurant food waste quantities from operational and
contextual features. A structured dataset was constructed by
integrating restaurant demand records, meteorological data and
temporal event indicators, yielding 77{,}980 records across 27
features. Because large-scale ground-truth food waste measurements
are not publicly available, the target variable was derived from
operationally justified assumptions, with the complete construction
formula and controlled stochastic variability disclosed for full
reproducibility. Four supervised regression models, namely Linear
Regression, Decision Tree, Random Forest and Gradient Boosting, were
evaluated under a chronological 70-30 train-test split that respects
the temporal ordering of restaurant operations, augmented by
5-fold time-series cross-validation. All reported metrics are
explicitly scoped to performance against the constructed target and
do not imply validation against measured food waste. Ensemble
methods consistently outperformed linear baselines. Random Forest
attained an MAE of 6.19\,kg, RMSE of 8.36\,kg and $R^{2}$ of 0.817
on the realistic feature subset following systematic exclusion of
algebraically leakage-prone variables. Feature importance analysis
identified menu diversity, operational area and temporal activity
patterns as the primary predictive drivers. The full dataset, target
construction formula, codebase and experimental configurations are
publicly released to support reproducibility and future extension
to empirically measured waste data.
\end{abstract}



\section{Introduction}

Food waste is a pressing global sustainability challenge with
far-reaching environmental, economic and social consequences.
The United Nations Environment Programme estimated that 1.05
billion tonnes of food were wasted at retail, food-service and
household levels in 2022, representing nearly one-fifth of the food
available to consumers \cite{ref1}. Food loss and waste are also
associated with approximately 8-10\% of annual global greenhouse
gas emissions and economic costs exceeding one trillion United
States dollars each year \cite{ref1}. Within hospitality operations,
overproduction, storage losses and plate waste remain important
sources of avoidable food waste \cite{ref7}. The food waste
hierarchy therefore prioritizes prevention before reuse, recycling,
recovery and disposal \cite{ref2}.

Advances in machine learning have created new opportunities for
data-driven demand forecasting, inventory optimization and
restaurant operations management \cite{ref3}. Random Forest is
well suited to modelling nonlinear relationships in structured
datasets \cite{ref21}. Gradient Boosting provides a complementary
sequential learning approach for complex predictive tasks
\cite{ref22}. XGBoost extends tree boosting through a scalable
system designed for efficient learning from structured data
\cite{ref25}. These techniques are relevant to restaurant waste
analysis, where operational scale, menu composition, promotional
activity and environmental factors can interact in complex ways.Despite this progress, publicly available datasets that combine
\emph{measured} restaurant food waste with rich operational and
contextual features remain scarce \cite{ref7}. Most existing
studies address food demand estimation rather than food waste
prediction directly, and few incorporate weather conditions,
holiday calendars or promotional signals as covariates. This
limits the generalizability of prior frameworks to real-world
restaurant waste management. The present work addresses these gaps through five contributions:

\begin{itemize}
  \item Construction of a structured, publicly available restaurant
        food waste dataset integrating operational, meteorological
        and temporal features, with the complete target construction
        formula and controlled stochastic variability disclosed for
        reproducibility.
  \item Systematic comparison of four supervised regression models
        under a chronological train-test split that respects
        operational time ordering, augmented by 5-fold time-series
        cross-validation.
  \item A rigorous feature leakage analysis and ablation study
        that identifies the primary leakage sources and quantifies
        their impact on reported performance.
  \item Identification of the operational and contextual drivers
        most strongly associated with the predicted waste target.
  \item A fully reproducible open-source framework to support
        future research incorporating empirically measured
        food waste data.
\end{itemize}

The remainder of this paper is structured as follows.
Section~\ref{sec:lit} reviews related work.
Section~\ref{sec:data} describes the dataset construction and
integration process. Section~\ref{sec:method} presents the
proposed methodology. Section~\ref{sec:results} reports
experimental results and discussion. Section~\ref{sec:conc}
concludes with directions for future work.


\section{Related Work}
\label{sec:lit}

Predictive analytics for food-service sustainability has attracted
growing research interest, encompassing food demand forecasting,
inventory optimization and waste reduction. Arunraj and Ahrens \cite{ref5} combined a seasonal autoregressive
integrated moving average model with quantile regression for daily
food sales forecasting. Their work demonstrates the operational
value of demand forecasting under variable sales conditions, but it
does not estimate restaurant food waste directly. Taylor \cite{ref8} developed an exponentially weighted quantile
regression approach for daily supermarket sales forecasting. The
study showed that robust demand modelling can improve point and
interval forecasts, although the target remained product sales
rather than food waste. Fildes et al.\ \cite{ref9} reviewed research and practice in retail
forecasting and highlighted the importance of data quality,
promotional information, evaluation design and operational context.
However, retail demand forecasting does not directly quantify
restaurant food waste. Carbonneau et al.\ \cite{ref10} investigated machine learning
techniques for supply-chain demand forecasting. Their findings
demonstrated the operational value of nonlinear predictive models,
but the framework focused on demand rather than direct waste
quantity estimation. Scherhaufer et al.\ \cite{ref11} quantified the environmental
impacts associated with food waste in Europe. Clowes et al.\
\cite{ref12} reported a strong business case for food waste
prevention in restaurant operations. Together, these studies
demonstrate the environmental and economic importance of improving
food waste management. Three methodological gaps remain across the literature. First,
publicly available datasets with \emph{directly measured} restaurant
food waste paired with operational and contextual features are
largely absent. Second, environmental and temporal covariates shown
to influence restaurant demand are infrequently included in waste
prediction models. Third, few studies conduct explicit feature
leakage analysis, which is critical when target variables are
partially constructed from operational proxies, and fewer still
employ chronological validation protocols suitable for time-ordered
operational data. The present work addresses all three gaps.


\section{Dataset Description and Integration}
\label{sec:data}

\subsection{Data Sources and Integration}

The dataset was constructed by integrating three categories of
publicly available information: restaurant demand records,
meteorological data and temporal event indicators. Restaurant
operational records provided daily identifiers, meal type, pricing,
operational area, promotional activity rates and customer visit
statistics. Meteorological data contributed daily temperature and
precipitation measurements aligned by date. Temporal indicators
encoded weekend, public holiday and special event status for each
record date.Integration was performed by aligning all sources on a composite
date and restaurant identifier key. Meteorological records were
matched to operational records by date, while temporal indicators
were merged using the same date key. The resulting integrated
dataset comprises 77{,}980 daily records across 27 features. After
integration, preprocessing steps, including missing-value
imputation, duplicate removal, ordinal label encoding of categorical
variables and standard scaling of continuous features, were applied
uniformly. The complete dataset, source code and reproducibility
resources used in this study are available in the
\href{https://github.com/mehedinaeem/ml-based-prediction-of-restaurant-food-waste-for-sustainable-food-management}
{project GitHub repository}.

\subsection{Target Variable Construction and Transparency}

Because no large-scale dataset of directly measured restaurant
food waste is publicly available, the target variable
\texttt{food\_waste\_kg} was estimated using the following
operationally grounded formula:

\begin{equation}
  W_i = \alpha \cdot P_i \cdot (1 - \beta \cdot C_i) + \epsilon_i
  \label{eq:target}
\end{equation}

\noindent where $W_i$ is the estimated daily food waste for record
$i$ in kilograms, $P_i$ is the estimated food prepared derived from
order volume and average meal weight, $C_i$ is the normalized
consumption ratio defined as sold meals relative to prepared meals,
$\alpha$ is a calibration scalar representing the assumed waste
proportion, $\beta$ is a consumption efficiency coefficient and
$\epsilon_i \sim \mathcal{N}(0,\sigma^2)$ is zero-mean Gaussian
noise introduced to reduce deterministic reconstruction of the
constructed target from any single observable feature. The parameter values $\alpha$, $\beta$ and $\sigma$, together with
the complete target construction code, are disclosed in the public
repository to support reproducibility and independent scrutiny. It
is essential to understand the scope of this design: models
evaluated against this target learn to approximate the constructed
formula with noise rather than ground-truth food waste. Reported
metrics such as $R^{2}=0.817$ quantify fit to the constructed target
and should not be interpreted as validated food waste predictions.
The framework is explicitly positioned as an exploratory
methodology study intended to establish a reproducible baseline for
future research incorporating empirically measured data.

\subsection{Feature Summary}

Table~\ref{tab:dataset_features} summarizes the key features
retained after leakage reduction for model development.

\begin{table}[htbp]
\caption{Summary of Key Dataset Features}
\label{tab:dataset_features}
\centering
\resizebox{\columnwidth}{!}{%
\begin{tabular}{ll}
\toprule
\textbf{Feature} & \textbf{Description} \\
\midrule
\texttt{restaurant\_id}          & Unique restaurant identifier \\
\texttt{city\_code}              & Geographic location code \\
\texttt{op\_area}                & Restaurant operational footprint \\
\texttt{unique\_meals}           & Distinct menu items served daily \\
\texttt{avg\_checkout\_price}    & Mean customer transaction value \\
\texttt{emailer\_promo\_rate}    & Email promotional campaign rate \\
\texttt{homepage\_feature\_rate} & Homepage promotional frequency \\
\texttt{is\_weekend}             & Weekend day indicator \\
\texttt{is\_holiday}             & Public holiday indicator \\
\texttt{temperature\_c}          & Daily mean temperature ($^\circ$C) \\
\texttt{food\_waste\_kg}         & Constructed daily food waste target \\
\bottomrule
\end{tabular}%
}
\end{table}


\section{Methodology}
\label{sec:method}

\begin{figure}[htbp]
\centering
\includegraphics[width=0.95\columnwidth]{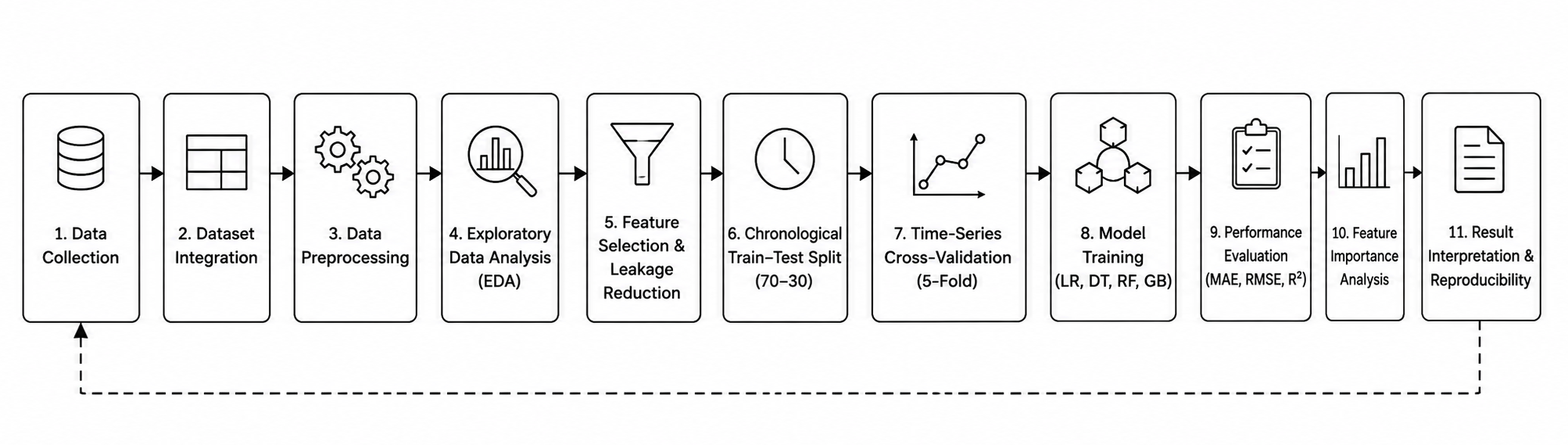}
\caption{Overall workflow of the proposed restaurant food waste
prediction framework.}
\label{fig_methodology}
\end{figure}

The proposed framework follows a structured pipeline comprising
data integration, preprocessing, exploratory analysis, feature
selection, model training and evaluation.
Figure~\ref{fig_methodology} illustrates the overall workflow.

\subsection{Data Preprocessing}

Preprocessing began with identification and removal of missing
values and duplicate records. Categorical variables including
restaurant category and cuisine type were encoded using ordinal
label encoding. Continuous operational and pricing features were
normalized through standard scaling with zero mean and unit
variance to reduce scale-induced bias and improve model convergence. Temporal features including year, month, ISO week number and
day-of-week indicator were extracted from the date attribute to
capture seasonal and cyclical patterns associated with restaurant
waste generation. These steps produced a consistent modelling
dataset while preserving the temporal variables required for
chronological evaluation.

\subsection{Exploratory Data Analysis}

\begin{figure}[htbp]
\centering
\includegraphics[width=0.9\columnwidth]{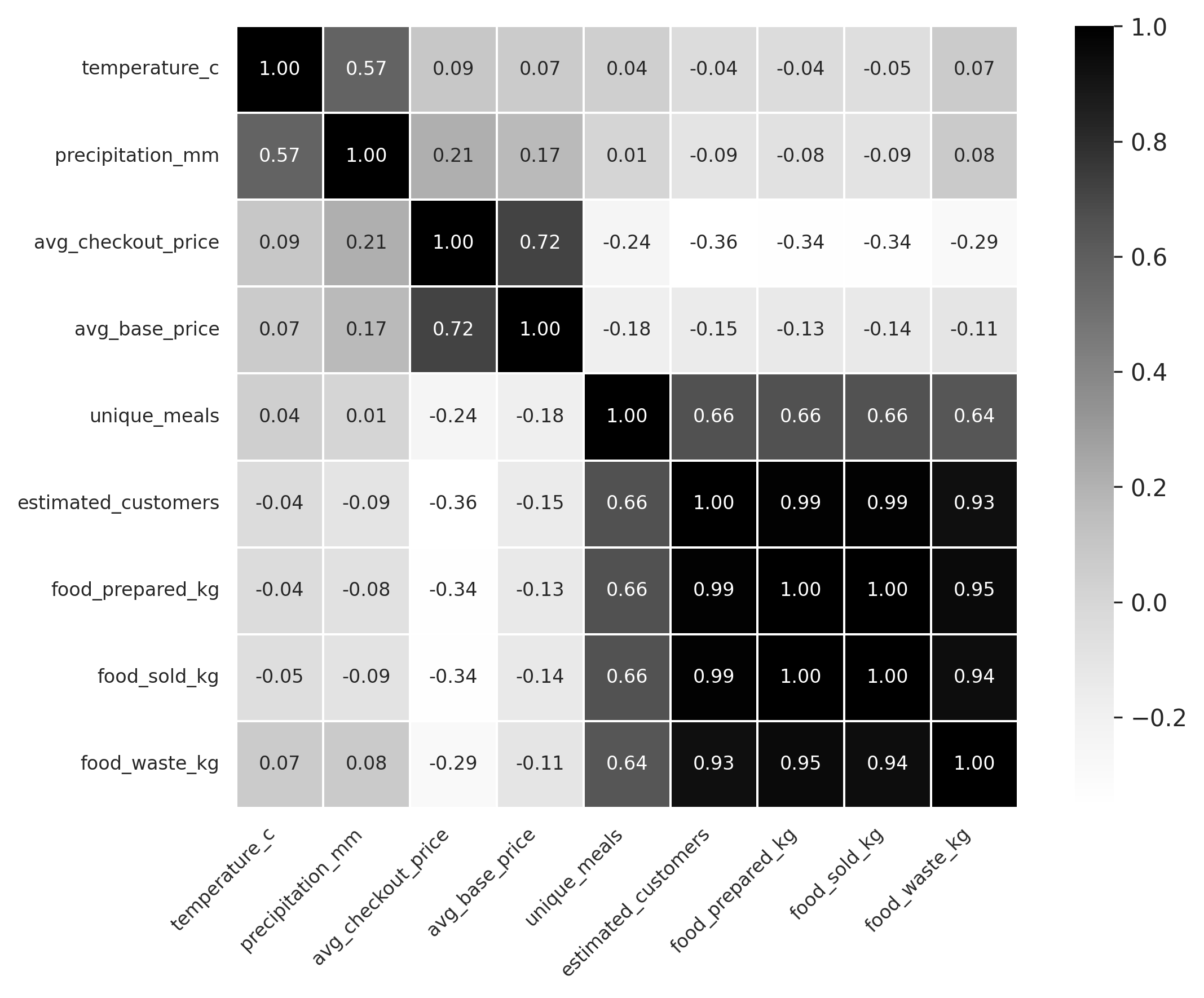}
\caption{Pearson correlation heatmap among primary numerical
features in the integrated dataset.}
\label{fig_correlation}
\end{figure}

Exploratory data analysis encompassed Pearson correlation analysis,
univariate distribution inspection, pairwise scatter visualization
and interquartile range based outlier detection.
Figure~\ref{fig_correlation} presents the correlation heatmap among
primary numerical features. Menu diversity
(\texttt{unique\_meals}) and operational area
(\texttt{op\_area}) exhibit the strongest positive associations
with the target variable, while temporal and contextual indicators
contribute moderate but consistent signal.

\subsection{Feature Selection and Leakage Reduction}

A key methodological challenge in working with operationally
constructed target variables is structural leakage. Features that
are algebraically entangled with the target derivation process can
yield unrealistically high performance scores that mask
generalization capability \cite{ref15}. Initial experiments using the full feature set including
\texttt{food\_prepared\_kg}, \texttt{food\_sold\_kg} and
\texttt{num\_orders} yielded $R^{2}$ values exceeding 0.99.
This result is expected because \texttt{food\_prepared\_kg} and
\texttt{food\_sold\_kg} participate directly in the target
construction represented by Equation~(\ref{eq:target}), allowing
models to reconstruct the target relationship rather than learn
broader predictive patterns. The variable \texttt{num\_orders} is
an algebraic antecedent of \texttt{food\_prepared\_kg} and
therefore constitutes indirect leakage. All three variables were
excluded from the \emph{realistic feature subset}, which retains
only variables plausibly available before service commencement:
restaurant size, cuisine category, pricing signals, promotional
indicators, weather conditions, holiday status and calendar
features.

\subsection{Machine Learning Models}

Four supervised regression models were evaluated:

\begin{itemize}
  \item \textbf{Linear Regression (LR):} Ordinary least-squares
        baseline providing an interpretable lower-bound reference.
  \item \textbf{Decision Tree (DT):} Nonparametric model capturing
        axis-aligned feature interactions, depth-limited at
        \texttt{max\_depth = 8} to control overfitting.
  \item \textbf{Random Forest (RF):} Bagging ensemble of 100
        decision trees with \texttt{max\_depth = 10}, exploiting
        variance reduction through bootstrap aggregation
        \cite{ref21}.
  \item \textbf{Gradient Boosting (GB):} Sequential residual-fitting
        ensemble with \texttt{n\_estimators = 100} and
        \texttt{learning\_rate = 0.05} \cite{ref22}.
\end{itemize}

Table~\ref{tab_hyperparameters} lists all configured
hyperparameters. All remaining settings retain Scikit-learn
defaults.

\begin{table}[htbp]
\caption{Hyperparameter Configuration of Evaluated Models}
\label{tab_hyperparameters}
\centering
\resizebox{\columnwidth}{!}{%
\begin{tabular}{ll}
\toprule
\textbf{Model} & \textbf{Key Hyperparameters} \\
\midrule
Linear Regression & Default OLS configuration with no regularization \\
Decision Tree & \texttt{max\_depth = 8} \\
Random Forest & \texttt{n\_estimators = 100}, \texttt{max\_depth = 10} \\
Gradient Boosting & \texttt{n\_estimators = 100}, \texttt{learning\_rate = 0.05} \\
\bottomrule
\end{tabular}%
}
\end{table}

\subsection{Chronological Train-Test Split and Cross-Validation}
\label{sec:split}

\begin{figure}[htbp]
\centering
\includegraphics[width=0.85\columnwidth]{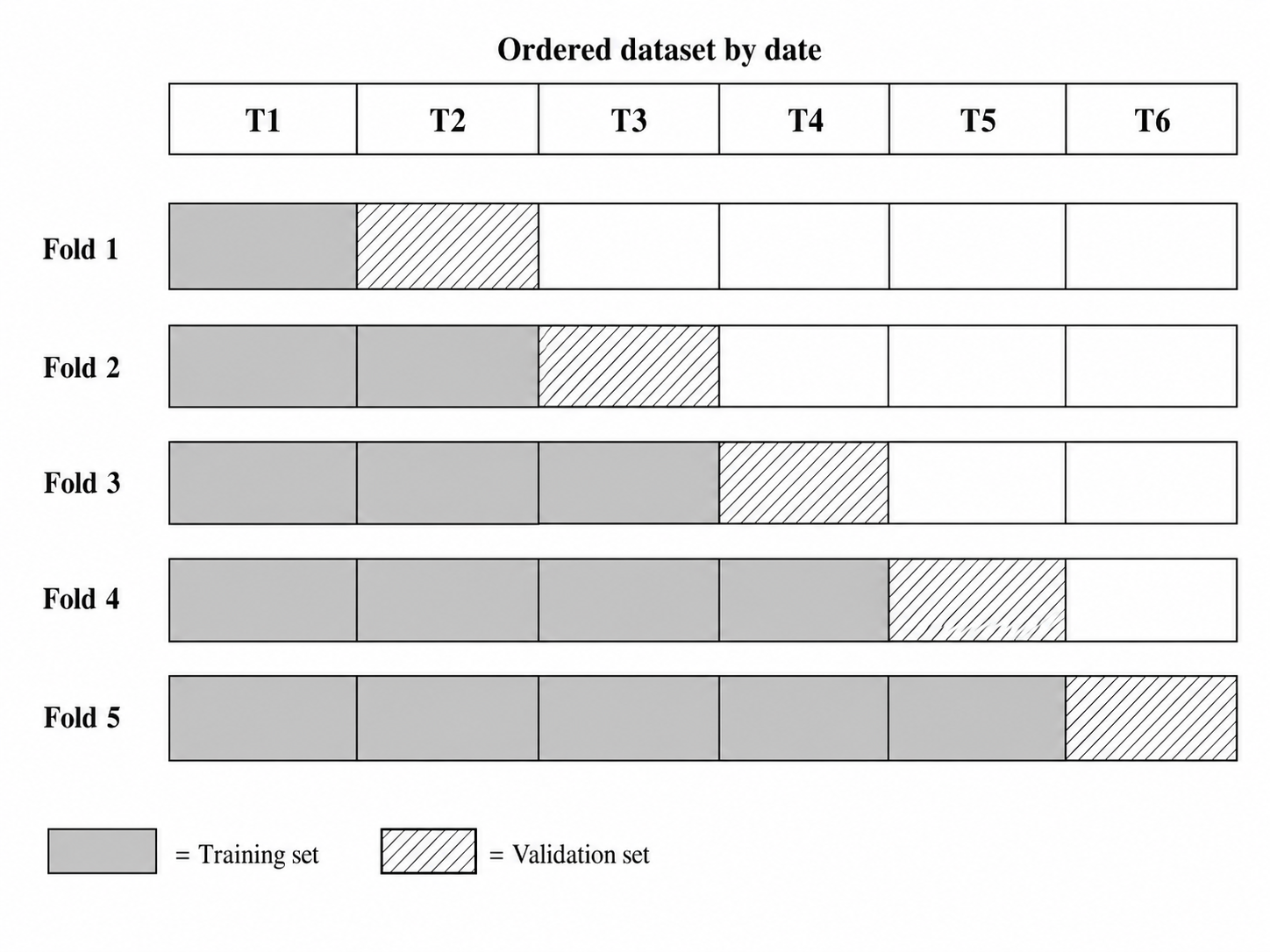}
\caption{Expanding-window 5-fold time-series cross-validation
protocol applied to the date-ordered training data.}
\label{fig_cv}
\end{figure}

Because the dataset represents time-ordered restaurant operations,
a random shuffle split is methodologically inappropriate because it
can allow future observations to inform predictions of past
records. The dataset was therefore sorted chronologically and split
at the 70th percentile of the date range, with the earlier 70\%
used for training and the remaining 30\% held out as the test set.
This preserves the temporal ordering of restaurant operations and
provides a more realistic evaluation of prospective forecasting
performance. To assess stability across different training windows, 5-fold
cross-validation was additionally applied using the
\texttt{TimeSeriesSplit} procedure. Each successive fold extends
the training set chronologically and evaluates the model on the
immediately following period. This expanding-window structure
follows established out-of-sample evaluation principles for
time-ordered data \cite{ref26}. Mean $R^{2}$ and standard deviation
across folds are reported alongside held-out test metrics.
Figure~\ref{fig_cv} illustrates the protocol.

\subsection{Evaluation Metrics}

Predictive performance was quantified using three complementary
regression metrics.

\subsubsection{Mean Absolute Error}

\begin{equation}
  \mathrm{MAE} =
  \frac{1}{n}\sum_{i=1}^{n}\left|y_i-\hat{y}_i\right|
\end{equation}

MAE reports the average magnitude of prediction errors in the
original unit of kilograms, providing an interpretable measure of
practical accuracy \cite{ref19}.

\subsubsection{Root Mean Squared Error}

\begin{equation}
  \mathrm{RMSE} =
  \sqrt{\frac{1}{n}\sum_{i=1}^{n}(y_i-\hat{y}_i)^2}
\end{equation}

RMSE penalizes large individual errors more heavily than MAE,
making it sensitive to outlying predictions \cite{ref19}.

\subsubsection{Coefficient of Determination}

\begin{equation}
  R^{2} =
  1-
  \frac{\displaystyle\sum_{i=1}^{n}(y_i-\hat{y}_i)^2}
       {\displaystyle\sum_{i=1}^{n}(y_i-\bar{y})^2}
\end{equation}

The coefficient of determination measures the proportion of target
variance explained by the model \cite{ref20}. Here $y_i$,
$\hat{y}_i$ and $\bar{y}$ denote the observed constructed value,
predicted value and mean target value, respectively. Given the
constructed nature of the target, $R^{2}$ should be interpreted as
a measure of model fit to the defined proxy rather than evidence of
real-world food waste prediction capability.


\section{Experimental Results and Discussion}
\label{sec:results}

All experiments were implemented in Python~3 using Scikit-learn,
Pandas and NumPy. The complete implementation, dataset, target
construction formula and reproducibility instructions are publicly
available through the project repository listed in
Section~\ref{sec:data}.

\subsection{Model Performance Under Chronological Split}

Table~\ref{tab_results} reports held-out test set performance for
all evaluated models on the realistic feature subset under the
chronological 70-30 split described in Section~\ref{sec:split}.

\begin{table}[htbp]
\caption{Test Set Performance on the Realistic Feature Subset}
\label{tab_results}
\centering
\small
\begin{tabular}{lccc}
\toprule
\textbf{Model} & \textbf{MAE (kg)} & \textbf{RMSE (kg)}
& \textbf{$R^{2}$\textsuperscript{\dag}} \\
\midrule
Linear Regression & 8.88 & 12.60 & 0.586 \\
Decision Tree & 7.50 & 10.24 & 0.726 \\
Gradient Boosting & 6.56 & 8.94 & 0.791 \\
Random Forest & \textbf{6.19} & \textbf{8.36}
& \textbf{0.817} \\
\bottomrule
\multicolumn{4}{l}{
\footnotesize
\textsuperscript{\dag}Metrics are evaluated against the constructed
proxy target.}
\end{tabular}
\end{table}

\begin{figure}[htbp]
\centering
\includegraphics[width=0.8\columnwidth]{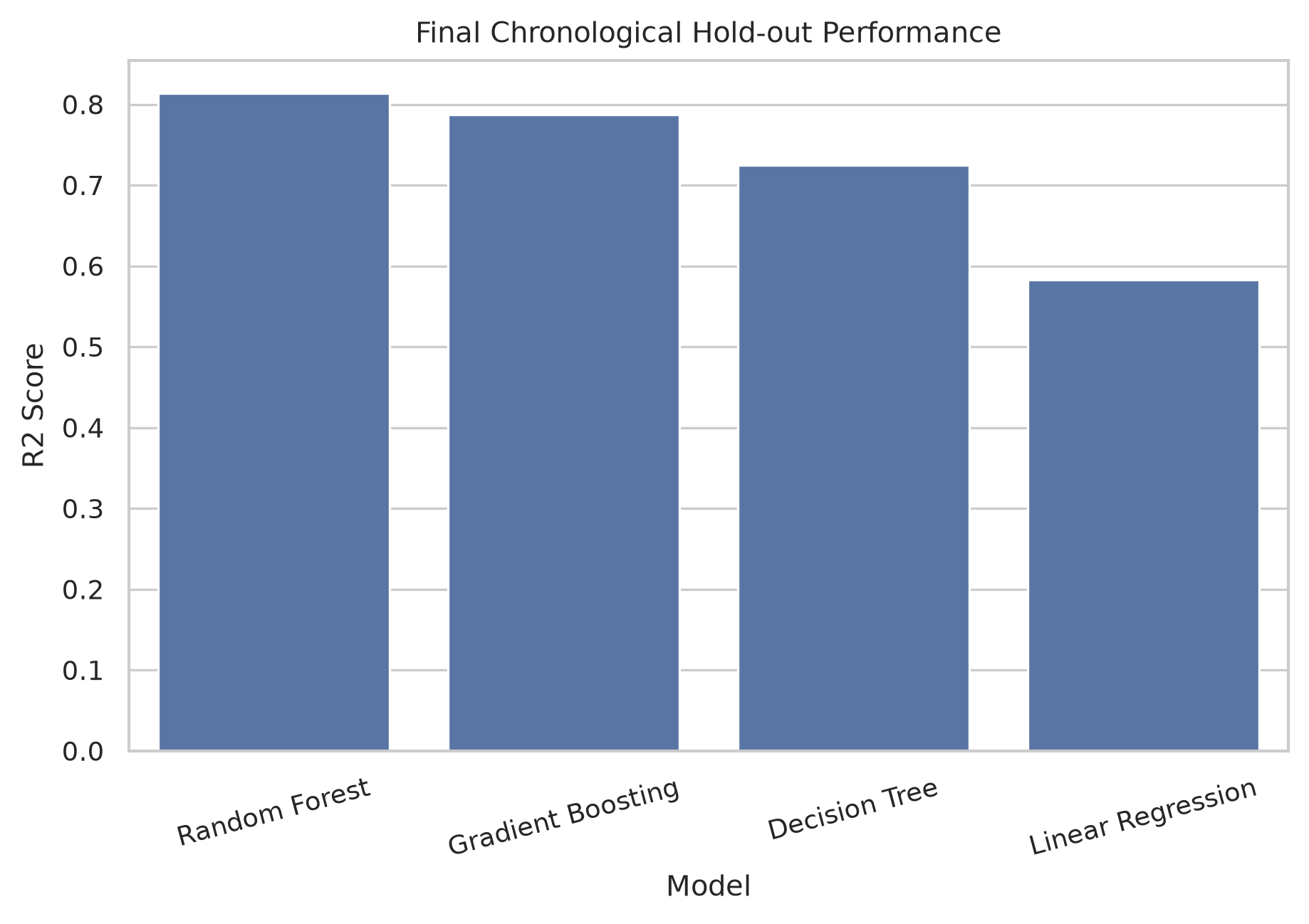}
\caption{$R^{2}$ score comparison of evaluated machine learning
models on the held-out chronological test set.}
\label{fig_model_comparison}
\end{figure}

Ensemble methods outperformed both baselines across all metrics.
Random Forest achieved the best result, reducing MAE by 30.3\%
relative to Linear Regression. The performance ordering
LR $<$ DT $<$ GB $<$ RF is consistent with the ability of bagging
ensembles to capture nonlinear interactions in structured
operational data \cite{ref21}. Gradient Boosting confirmed the
complementary strength of sequential residual-fitting approaches
\cite{ref22}. Figure~\ref{fig_model_comparison} provides a visual
summary.

\subsection{Cross-Validation Results}

Table~\ref{tab_cv_results} presents mean $R^{2}$ and standard
deviation across the five time-series cross-validation folds.

\begin{table}[htbp]
\caption{5-Fold Time-Series Cross-Validation Results}
\label{tab_cv_results}
\centering
\begin{tabular}{lcc}
\toprule
\textbf{Model} & \textbf{Mean $R^{2}$} & \textbf{Std.} \\
\midrule
Linear Regression & -0.033 & 0.649 \\
Decision Tree & 0.595 & 0.091 \\
Gradient Boosting & 0.716 & \textbf{0.032} \\
Random Forest & \textbf{0.720} & 0.072 \\
\bottomrule
\end{tabular}
\end{table}

Random Forest achieved the highest mean cross-validation $R^{2}$
of 0.720, while Gradient Boosting showed the lowest variability
among the nonlinear models with a standard deviation of 0.032.
The held-out test performance of Random Forest was higher than its
cross-validation mean, indicating that predictive performance
varies across successive temporal windows.

\subsection{Feature Importance Analysis}

\begin{figure}[htbp]
\centering
\includegraphics[width=0.90\columnwidth]{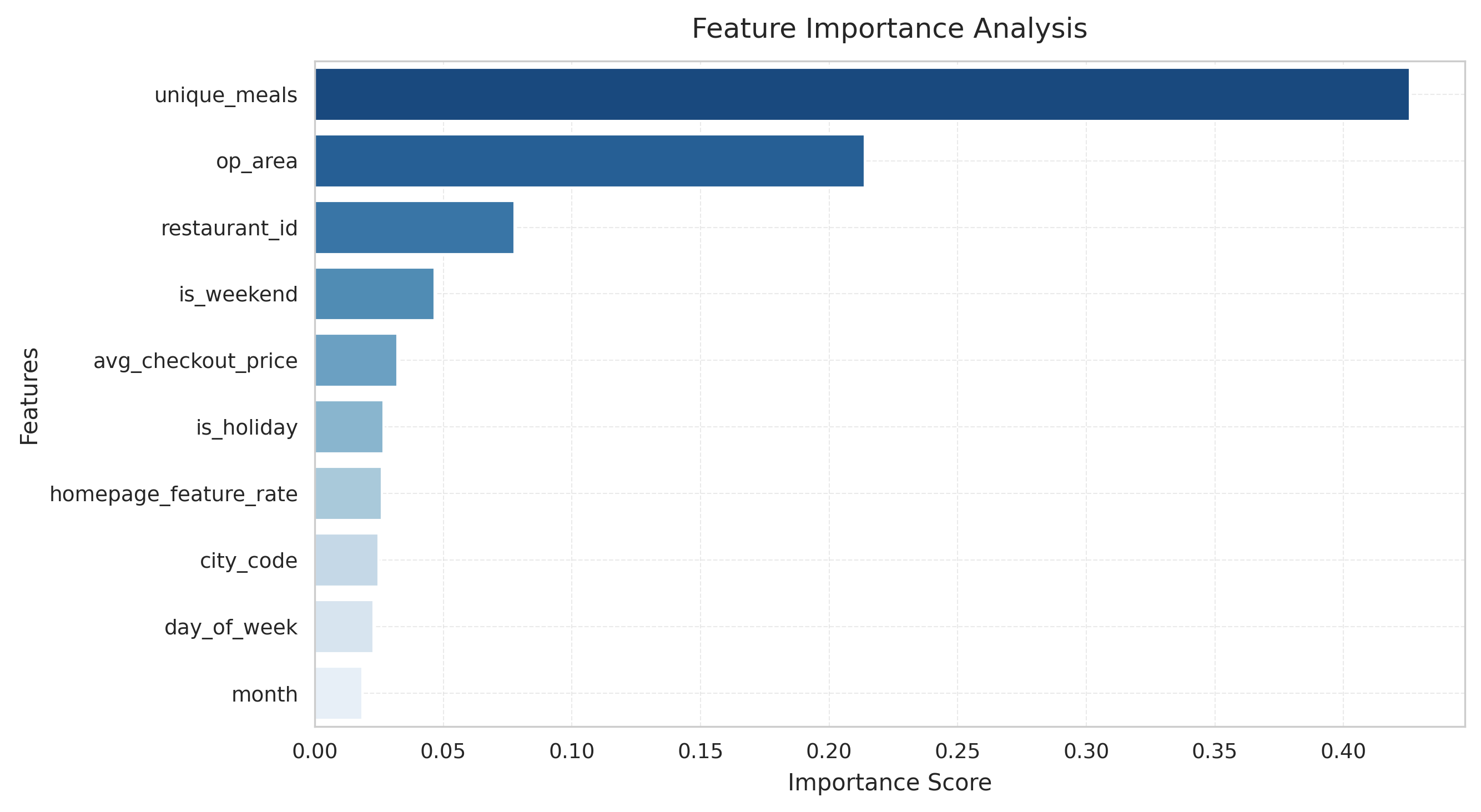}
\caption{Random Forest feature importance scores for the realistic
feature subset.}
\label{fig_feature_importance}
\end{figure}

Feature importance scores were derived from the impurity-based mean
decrease in node impurity accumulated across all trees of the
Random Forest model. Figure~\ref{fig_feature_importance} presents
the ranked importance profile for the realistic feature subset.

Menu diversity represented by \texttt{unique\_meals} and
operational footprint represented by \texttt{op\_area} emerged as
the two most influential features. This is operationally
consistent because restaurants offering broader menus and
operating larger facilities are expected to prepare more food and
generate proportionally more waste. Weekend and holiday indicators
provided moderate contributions, reflecting demand changes during
high-traffic periods. Pricing and promotional features contributed
smaller but non-negligible importance. These variables nevertheless
contributed less than menu diversity and operational area.

\subsection{Ablation and Leakage Analysis}

Table~\ref{tab_ablation} summarizes the impact of progressive
feature exclusion on Random Forest performance. This experiment
directly identifies which feature groups constitute structural
leakage.

\begin{table}[htbp]
\caption{Ablation Study: Effect of Feature Selection on $R^{2}$}
\label{tab_ablation}
\centering
\small
\begin{tabular}{p{0.79\columnwidth}c}
\toprule
\textbf{Feature Configuration} & \textbf{$R^{2}$} \\
\midrule
All features including \texttt{food\_prepared\_kg},
\texttt{food\_sold\_kg} and \texttt{num\_orders} & 0.997 \\
\texttt{food\_prepared\_kg} and \texttt{food\_sold\_kg}
removed & 0.990 \\
\texttt{num\_orders} additionally removed, producing the realistic
subset & 0.817 \\
\bottomrule
\end{tabular}
\end{table}

The results clarify the leakage structure. Removing
\texttt{food\_prepared\_kg} and \texttt{food\_sold\_kg}, which
participate directly in the target construction represented by
Equation~(\ref{eq:target}), reduces $R^{2}$ by 0.007. This
indicates that \texttt{num\_orders} remains a substantial indirect
leakage source because it encodes the order volume from which
\texttt{food\_prepared\_kg} is derived. The model can therefore
reconstruct part of the target relationship indirectly. Removing
\texttt{num\_orders} reduces $R^{2}$ to 0.817, representing
performance on the reduced feature subset containing variables
available before service commencement. This stepwise analysis
provides a transparent account of how target-related variables
influence the reported performance.

\begin{figure}[htbp]
\centering
\includegraphics[width=0.86\columnwidth]{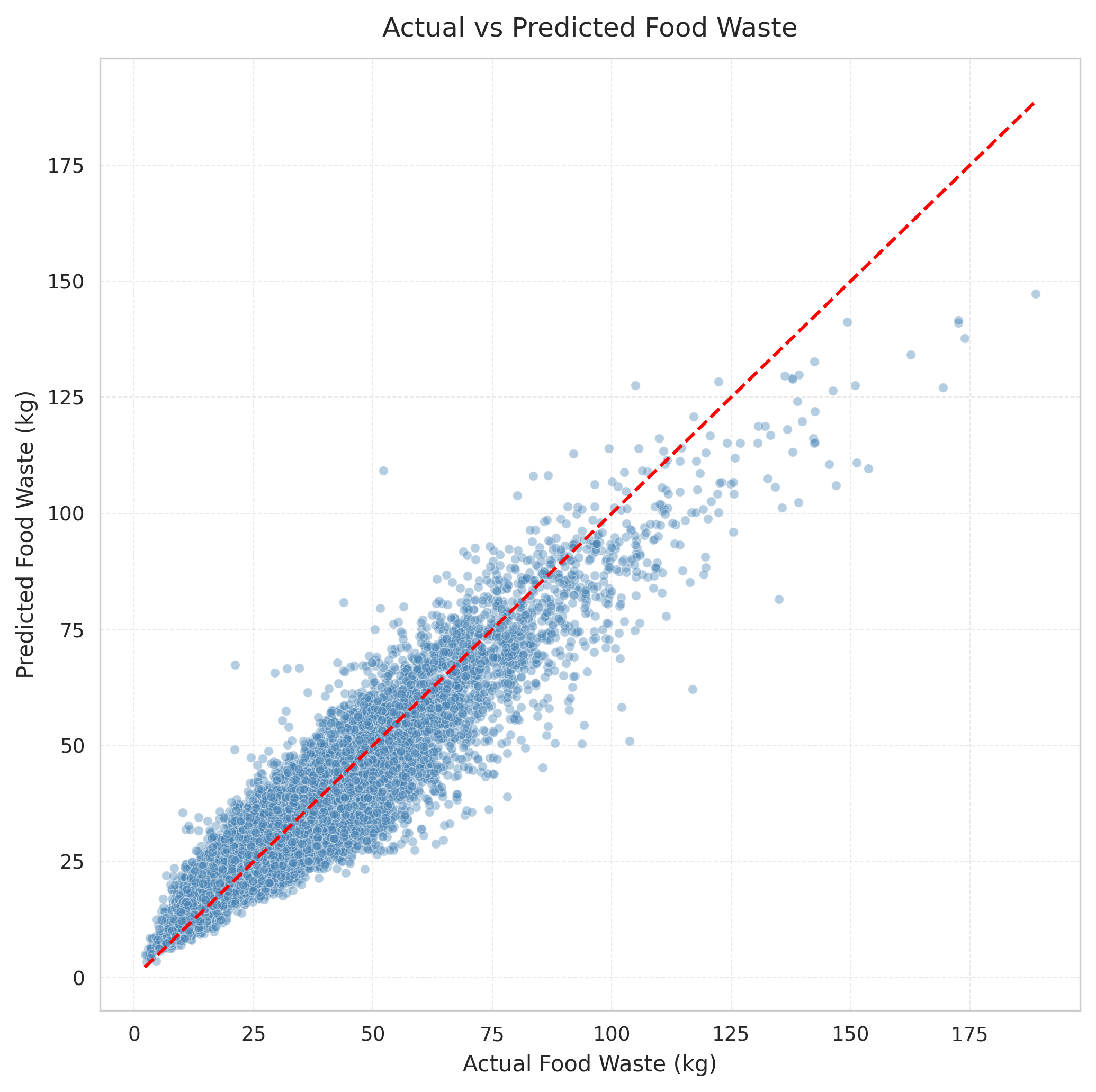}
\caption{Constructed target versus predicted food waste in
kilograms for Random Forest on the chronological held-out test
set.}
\label{fig_actual_predicted}
\end{figure}

Figure~\ref{fig_actual_predicted} visualizes the alignment between
constructed target values and Random Forest predictions on the
chronological test set.

\subsection{Discussion}

The experimental results demonstrate that the realistic feature
subset comprising variables available before service commencement
contains meaningful predictive signal for the constructed food
waste target. Ensemble learning methods were effective at
capturing nonlinear interactions among operational, contextual and
temporal variables.

Three scope limitations of the present study must be clearly
stated. First, all reported metrics quantify model fit against a
proxy target constructed by the authors and do not constitute
validation against empirical food waste measurements. Second, the
chronological split and time-series cross-validation adopted in
this work provide temporally ordered evaluation, but walk-forward
validation using unseen restaurant locations would provide a
stronger test of temporal and geographic generalizability. Third,
although Gaussian noise reduces exact deterministic reconstruction,
some residual correlation between the noise-free component of the
target and the realistic feature subset may persist, particularly
through features such as \texttt{unique\_meals} that partially
inform \texttt{food\_prepared\_kg}. Quantifying this residual
relationship through partial dependence analysis and other
explainability methods is an important direction for future work.

From a deployment perspective, the features identified as most
important, including menu diversity, operational area and temporal
indicators, are available before service commencement. This makes
the framework applicable in principle to next-shift waste
forecasting without real-time instrumentation. Deployment in live
restaurant environments would require integration with
point-of-sale systems, kitchen management platforms and direct
waste monitoring hardware to replace the proxy-derived target with
empirically measured food waste quantities.


\section{Conclusion}
\label{sec:conc}

This paper presented an exploratory machine learning framework for
estimating daily restaurant food waste from operational,
meteorological and temporal features. A structured dataset of
77{,}980 records was assembled from publicly available sources, and
a methodologically transparent proxy target was constructed using
an explicit formula with controlled Gaussian noise. The target
construction process and assumptions are disclosed to support
reproducibility. Four supervised regression models were evaluated under a
chronological 70-30 train-test split and a 5-fold time-series
cross-validation protocol. Ensemble methods consistently
outperformed linear baselines. Random Forest achieved
MAE$=6.19$\,kg, RMSE$=8.36$\,kg and $R^{2}=0.817$ against the
constructed proxy target on the realistic feature subset. A
stepwise ablation and leakage analysis identified
\texttt{num\_orders} as the primary residual leakage source, and
its exclusion produced the reported realistic feature subset.
Feature importance analysis identified menu diversity and
operational area as the dominant predictive drivers. The full codebase, dataset, target construction formula and
evaluation scripts are publicly released to support independent
reproducibility and community-driven extension. Future work will
pursue three directions: collection and integration of directly
measured restaurant food waste to replace the proxy target and
validate the framework against ground truth; evaluation of advanced
gradient-boosted methods including XGBoost \cite{ref25}, LightGBM
\cite{ref27} and CatBoost \cite{ref28}; and development of
walk-forward, out-of-sample cross-restaurant validation protocols
to rigorously assess temporal and geographic generalizability.


\bibliographystyle{IEEEtran}
\bibliography{biblio}

\end{document}